\documentclass[journal,twocolumn]{IEEEtran}

\usepackage{amsmath}
\usepackage{caption}
\usepackage{algorithm}
\usepackage{algpseudocode}
\usepackage{setspace}
\usepackage{color,hyperref}
\usepackage{subcaption}
\usepackage{graphicx}
\usepackage{amsmath,amssymb} 
\usepackage{epsfig}
\usepackage{amsmath}
\usepackage{amssymb}
\usepackage{colortbl}
\usepackage{soul}
\usepackage{multirow}
\usepackage{pifont}

\definecolor{mygray}{gray}{.9}

\newlength\savedwidth
\newcommand\whline{\noalign{\global\savedwidth\arrayrulewidth
		\global\arrayrulewidth 1.25pt}%
	\hline
	\noalign{\global\arrayrulewidth\savedwidth}}

\ifCLASSINFOpdf
\else
\fi

\definecolor{darkblue}{rgb}{0.0,0.0,1.0}
\hypersetup{colorlinks,breaklinks,
	linkcolor=darkblue,urlcolor=darkblue,
	anchorcolor=darkblue,citecolor=darkblue}

\begin{document}
\setul{}{1.5pt}

\title{PCC Net: Perspective Crowd Counting via Spatial Convolutional Network}

\author{Junyu~Gao, Qi~Wang,\IEEEmembership{~Senior Member,~IEEE}, and ~Xuelong~Li,\IEEEmembership{~Fellow,~IEEE}
	\thanks{
	
	This work was supported by the National Natural Science Foundation of China under Grant U1864204 and 61773316, Natural Science Foundation of Shaanxi Province under Grant 2018KJXX-024, and Project of Special Zone for National Defense Science and Technology Innovation. (Corresponding author: Qi Wang.)	
	
	Junyu Gao, Qi Wang and Xuelong Li are with the School of Computer Science and the Center for OPTical IMagery Analysis and Learning (OPTIMAL), Northwestern Polytechnical University, Xi'an 710072, China (e-mail: gjy3035@gmail.com; crabwq@gmail.com; li@nwpu.edu.cn).
		
	Copyright \copyright 20xx IEEE. Personal use of this material is permitted. However, permission to use this material for any other purposes must be obtained from the IEEE by sending an email to pubs-permissions@ieee.org.
}
}
\markboth{{IEEE} Transactions on Circuits and Systems for Video Technology}%
{Shell \MakeLowercase{\textit{et al.}}: Bare Demo of IEEEtran.cls for Journals}
\maketitle

\begin{abstract}
Crowd counting from single image is a challenging task due to high appearance similarity, perspective changes and severe congestion. Many methods only focus on the local appearance features and they cannot handle the aforementioned challenges. In order to tackle them, we propose a Perspective Crowd Counting Network (PCC Net), which consists of three parts: 1) Density Map Estimation (DME) focuses on learning very local features for density map estimation; 2) Random High-level Density Classification (R-HDC) extracts global features to predict the coarse density labels of random patches in images; 3) Fore-/Background Segmentation (FBS) encodes mid-level features to segments the foreground and background. Besides, the DULR module is embedded in PCC Net to encode the perspective changes on four directions (Down, Up, Left and Right). The proposed PCC Net is verified on five mainstream datasets, which achieves the state-of-the-art performance on the one and attains the competitive results on other four datasets. The source code are available at \url{https://github.com/gjy3035/PCC-Net}.

\end{abstract}

\section{Introduction}
\label{Sec_intro}

\begin{figure*}
	\centering
	\includegraphics[width=0.75\textwidth]{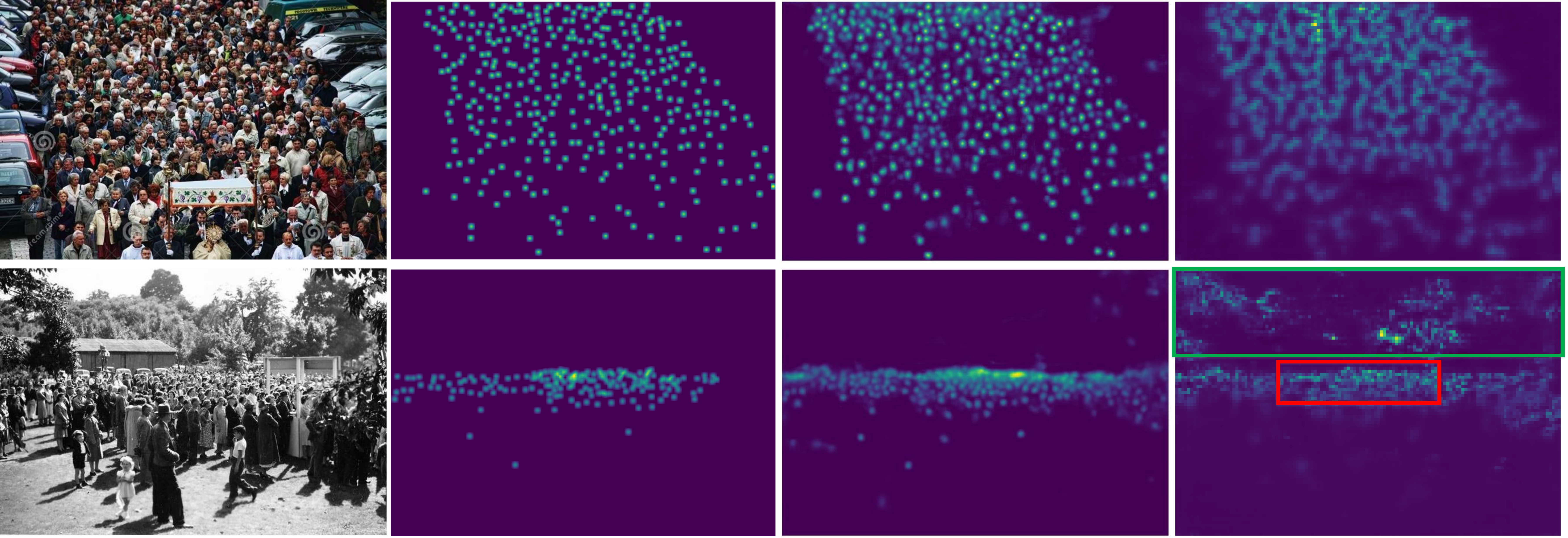}
	\caption{The comparison results of the proposed PCC Net and the state-of-the-art method \cite{li2018csrnet}. First Column: original image; Second Column: ground truth; Third Column: the results of PCC Net; Last Column: CSRNet \cite{li2018csrnet}. }\label{Fig-intro}
\end{figure*}

Crowd analysis is a hot topic in the computer vision because of its strong applied value in many areas: video surveillance, public safety, urban planning, behavior understanding and so on \cite{7801078,7509595,chen2017patch,li2017multiview,yu2017multitask}. In this paper, we dedicate to the crowd counting task which generates a density map and predicts the number of people for the given crowd scenes. Fig. \mbox{\ref{Fig-intro}} intuitively describes the density map estimation. The number of people in the scene is the sum of all pixels's values.

Recently, with the development of deep learning, many CNN-based methods \cite{zhang2015cross,onoro2016towards,sam2017switching,sindagi2017generating,8360001} achieve the amazing improvements for crowd counting compared with traditional methods \cite{idrees2013multi,chen2013cumulative,lempitsky2010learning}. The survey \mbox{\cite{sindagi2018survey}} further analyzes the CNN-based crowd counting methods. However, because of high clutter, high appearance similarity and complex perspective changes, the above methods cannot perform well. At present, there are two intractable problems in the crowd counting field: 1) some background regions are similar to the congested region, which is usually prone to misestimation of the density; 2) the perspective change in crowd scenes cannot be effectively encoded, which causes the poor quality of density map. For the first problem, many methods \cite{idrees2013multi,wang2016fast,liu2017decidenet} attempt to detect each head in the crowd scenes. However, the heads in the high density area are too tiny to be detected effectively and accurately. As for the second problem, some methods \cite{marsden2016fully,sindagi2017cnn} process the whole image and exploit the global contextual information. However, these methods do not effectively encode the complex perspective changes. 

Here, we further explain the perspective changes. For a common crowd scene, the density distribution in physical world does not suffer from perspective. However, in a 2-D image, the population density increases as the distance of the scene is getting farther, of which the most fundamental reason is perspective phenomenon. Note that the population density means the region can accommodate the density of people, which is not same as crowd density. Thus, in a single image, the most intuitive embodiment is a trend of population density caused by perspective changes.

In order to tackle the above problems, in this paper, we propose a multi-task Perspective Crowd Counting Network (PCC Net) to mine the global and perspective information in the crowd scenes. On one hand, we present the multi-task learning to segment the head and the background region. Compared with the detection-based method \cite{idrees2013multi,wang2016fast,liu2017decidenet}, our treatment segments the high congested region better. On the other hand, we design a perspective module (called as ``DULR module'') to encode the perspective changes on four directions, namely Down, Up, towards the Left and Right. Especially, for the congested crowd scenes, the proposed method effectively encodes the perspective changes and accurately estimates the density. 

In addition, motivated by a multi-task learning method \mbox{\cite{sindagi2017cnn}}, we design a more flexible strategy to extract the high-level features. To be specific, during the training stage, whole images are fed to the network. Then some random regions' feature maps are extracted to be fed into a classification network. All regions are assigned with coarse density labels. Compared with the previous method \mbox{\cite{sindagi2017cnn}} that generates the training set in advance, the training set generated by our strategy is more diversified than the fixed training set in \mbox{\cite{sindagi2017cnn}}.

Fig. \mbox{\ref{Fig-intro}} shows some challenging exemplars and the results of the proposed PCC Net and the state-of-the-art method \mbox{\cite{li2018csrnet}}. As for the first input with uniform perspective, our result is close to the ground truth while the CSRNet's result is inconsistent with the ground truth. For the second row, the CSRNet ignores the perspective changes so that the density in congested region increases steeply (as shown in the red box). However, PCC Net's output shows the similar perspective density trends with the ground truth. Besides, CSRNet mistakenly estimates the background trees in green box. In general, the proposed PCC Net shows significantly improvements than CSRNet and effectively alleviates the aforementioned problems.

\begin{figure*}[t]
	\centering
	\includegraphics[width=0.98\textwidth]{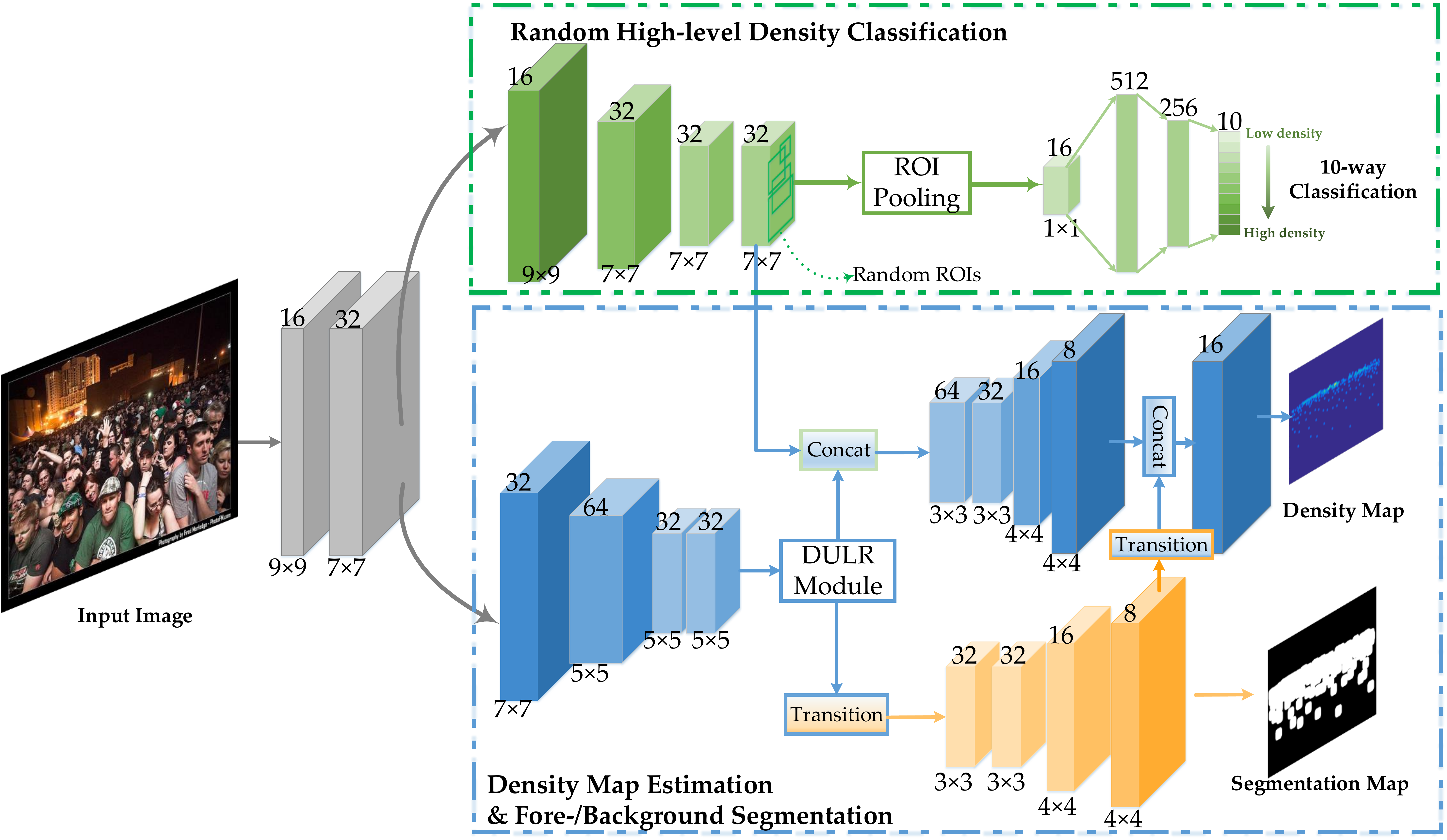}
	\caption{Overview of the proposed PCC Net architecture. It consists of Density Map Estimation (DME), Random High-level Density Classification (R-HDC) and Fore-/Background Segmentation (FBS).}\label{Fig-overview}
\end{figure*}

The overview of our method is described below. The proposed PCC Net is composed of three tasks: Density Map Estimation (DME), Random High-level Density Classification (R-HDC), and Fore-/Background Segmentation (FBS). To be specific, R-HDC is regarded as image-level classification, and DME, FBS are treated as pixel-level regression and classification, respectively. Note that the perspective module is added to the pixel-level streams. Fig. \ref{Fig-overview} shows the architecture of PCC Net. 

In summary, the main contributions of this paper are:
\begin{enumerate}
	\item[1)]Propose a Random High-level Density Classification (R-HDC) to obtain more diversified training samples for density classification, which is more flexible than the previous methods and obtain a better performance. 
	
	\item[2)]Present the Fore-/Background Segmentation (FBS) to extract region-level features to segment fore- and background. It significantly alleviates some erroneous estimations for background regions.
	
	\item[3)]Design a perspective DULR module based on spatial convolutional networks to encode the perspective changes. As for the extremely congested area, the perspective module can effectively improve the quality of density map.

\end{enumerate}

\section{Related Work}
In this section, we review important CNN-based crowd counting
and density estimation methods. In addition, the proposed PCC Net exploits the segmentation and spatial CNN for crowd counting, so some related works about them are briefly described.

\textbf{Patch-based crowd counting.} 
Since patch-based methods can effectively capture local features and generate a great deal of training data, many methods \cite{fu2015fast,wang2015deep,zhang2015cross,sam2017switching,8360001} crop the images with different sizes to train the model, and predict each sliding window during the testing phase. In 2015, Convolutional Neural Networks (CNN) is applied in crowd counting by Fu \emph{et al.} \cite{fu2015fast} and Wang \emph{et al.} \cite{wang2015deep}, which are simple single-model methods. Fu \emph{et al.} \cite{fu2015fast} classify the image into one of the five classes according the density instead of directly estimating density maps. Wang \emph{et al.} \cite{wang2015deep} propose an end-to-end CNN regression model for counting people of images in extremely dense crowds, which is finetuned on the AlexNet. However, the above single-model methods perform poorly for unseen target crowd scenes. To handle it, Zhang \emph{et al.} \cite{zhang2015cross} propose a data-driven method to finetune the trained CNN model on the target scene. In addition, they adopt a multi-task scheme to predict crowd density and crowd count with two related learning objectives. Sam \emph{et al.} \cite{sam2017switching} proposed a switching CNN that automatically selects an optimal regressor from several independent regressors for each input patch. Wang \emph{et al.} \cite{7927432} propose a deep metric learning to extract the local structural information. Kang \emph{et al.} \cite{8360001} further compare crowd density maps that generated by some mainstream methods, on some crowd analysis tasks, including crowd counting, object detection, and object tracking in the filed of video surveillance.

\textbf{Whole image-based crowd counting.} 
Since the patch-based methods cannot encode the global contextual information, some works \cite{zhang2016single,shang2016end,sindagi2017cnn,sindagi2017generating} focus on the whole image-based scheme. Zhang \emph{et al.} \cite{zhang2016single} propose a Multi-column CNN to predict density map, which processes the input image with arbitrary size or resolution. Shang \emph{et al.} \cite{shang2016end} propose an end-to-end network that consists of CNN model and Long-short time memory (LSTM) decoder to predict the number of people. Sheng \emph{et al.} \cite{sheng2016crowd} present a novel image representation which takes into consideration semantic attributes and spatial cues. Sindagi \emph{et al.} \cite{sindagi2017cnn} propose a cascaded CNNs framework to incorporate learning of a high-level prior to prompt the quality of the density map. Sindagi and Vishal \cite{sindagi2017generating} propose a Contextual Pyramid CNN to generate the high-quality density map, which encodes the image at the patch and whole image level. To be specific, 
it contains three estimators: Global Context Estimator, Local Context Estimator and Density Map Estimator. The first two estimators generate the contextual information on the patches, and the last estimator outputs the density map of the whole image. 

\textbf{Image segmentation and spatial CNN.} In 2015, Long \emph{et al.} \cite{long2015fully} propose a Fully Convolutional Networks (FCN) for semantic segmentation, which is a variant of traditional CNN. Gao \emph{et al.} \cite{gao2017conf_embedding} propose a Siamesed FCN (s-FCN) to combine RGB and contour information for object segmentation. In order to encode the contextual and spatial information, some novel CNN architectures \cite{yu2015multi,liu2017learning,pan2017spatial,gao2018jounal_joint} design the spatial propagation operation. Yu and Vladlen \cite{yu2015multi} develop a dilated convolution to systematically aggregate multi-scale contextual information without losing resolution. Liu \emph{et al.} \cite{liu2017learning} present a Spatial Propagation Networks (SPN) for learning the affinity matrix for vision tasks, which is a three-way connection for the linear propagation model. Pan \emph{et al.} \cite{pan2017spatial} design a slice-by-slice convolutional operation within feature maps, which propagates spatial information across rows and columns in a layer. Wang \emph{et al.} \cite{gao2018jounal_joint} present a soft restricted Markov Random Field (MRF) to encode the spatial information in the urban scenes, which effectively remedy the over-smoothness phenomenon in second-order term of traditional MRF.

\section{Method}

The proposed PCC Net consists of three tasks: Density Map Estimation (DME), Random High-level Density Classification (R-HDC) and Fore-/Background Segmentation (FBS). R-HDC randomly crops the given images and classifies them as high-level labels. DME and FBS are Fully Convolutional Network (FCN): DME generates the crowd density maps, and FBS segments the head regions and background. In addition, the DULR module is added to the FCN to encode the perspective changes.

\subsection{Density Map Estimation}
\label{Sec_DME}

Density map estimation (DME) and semantic segmentation are pixel-wise problems: regression and classification, respectively. Thus, many works in crowd counting adopt the theories in semantic segmentation. In 2015, Long \emph{et al.} \cite{long2015fully} propose a Fully Convolutional Network (FCN) to focus on pixel-wise classification. Compared with traditional CNN, FCN removes the fully connected layers, so it can take input of arbitrary size. Besides, FCN adds the deconvolutional layers to upsample the feature map. Thus, it can generate the correspondingly-sized output to classify an image at the pixel level.

In the field of crowd counting, some current methods remove the classification layer from the original FCN to tackle the density map estimation and achieve the significant improvements. In this paper, we design a simple and effective FCN to predict the crowd density map. The blue blocks in Fig. \ref{Fig-overview} show the feature map flow of the proposed DME. And the numbers on the top and bottom of the block respectively denote the channel and filter size.

During the training phase, the loss function is standard Mean Squared Error (MSE).

\subsection{Random High-level Density Classification}
\label{Sec_R-HDC} 
FCN-based density map estimation is a pixel-wise regression task, which focuses on local feature extraction. However, it ignores the global contextual information, which is important to alleviate mistake estimation. To this end, \cite{sindagi2017cnn} propose a multi-task networks to estimate the density map and predict the high-level density labels. It quantizes the density into ten categories at the image level as high-level labels. However, the learning strategy is not elegant: the training samples are image patches from the original dataset, and they are fixed during the training stage, which impacts the classification performance for the original datasets.

In this paper, inspired by the method \cite{7410526}, we design a more flexible end-to-end training scheme, which is called as ``Random High-level Density Classification (R-HDC)''.

During the training process, it can process the whole images and generate the infinite amount of images patches to cover the original datasets. For each image, the certain number of Regions of Interest (ROIs) are generated randomly. In order to learn effective global features, the ROIs' sizes are so large that they can cover more than $1/16$ of the whole image. The ROIs' feature maps are obtained by the ROI pooling operation \cite{7410526}. Then fully connected networks assign the high-level density labels for the ROIs. The proposed R-HDC cover the inputs as many as possible, which is treated as a data augmentation method. In addition, training with the whole images allows networks to encode global features. The detailed algorithm is shown in Section \ref{Sec_DULR}. 

In practice, R-HDC is a 10-class classification problem. We compute the density of each image in the dataset, and divide it into ten levels according to the density. Here, the density is defined as the ratio of people number to the area of image (namely $height \times width$). During the training phase, for each ROI, we online compute its high-level label and make the model to learn to predict it.

In order to train R-HDC, the objective is minimizing the standard cross-entropy loss.

\subsection{Fore-/Background Segmentation}

\label{Sec_FBS}

The above multi-task framework (DME+R-HDC) extracts the local and global features to predict more accurate density map than the single DME. However, during the generation process of density map, the kernel size is smaller than head size in many cases. Therefore, DME+R-HDC neglects the structural shape feature in sparse crowd region and the contextual information in congested crowd region. 

\begin{figure}[htbp]
	\centering
	\includegraphics[width=0.48\textwidth]{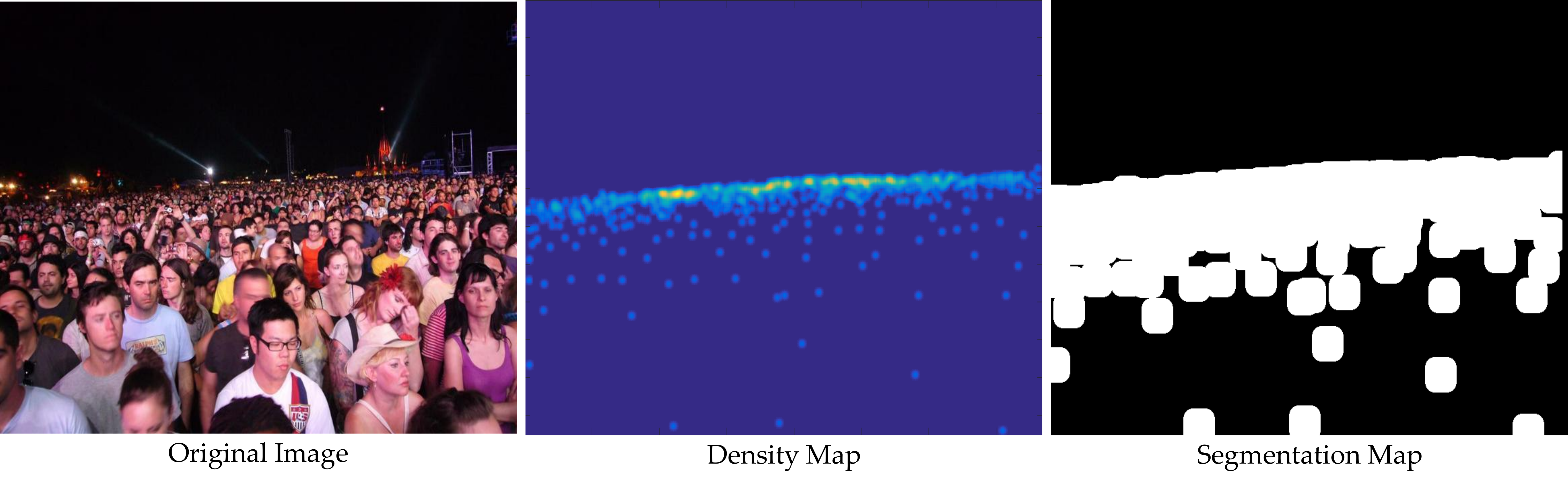}
	\caption{The comparison between the density map and the segmentation map.}\label{Fig-Map}
\end{figure}

In order to reduce the above problem, we attempt a head segmentation method to learn more large-scale distinguished features. Compared with density maps, the head region is larger than the heat region in density map, which means that the former covers more contextual information. Fig. \mbox{\ref{Fig-Map}} intuitively shows the comparison between the density map and segmentation map. From it, the segmentation map covers more large-range head area than the density map. The larger region contains the whole head shape, the face structure and the crowd distribution. The high-level information aids the model to learn more semantic features.

Unfortunately, annotating foreground region consumes a lot of labor and resources. Thus, we exploit the key point provided by the dataset to generate the coarse segmentation labels. To be specific, the binary image is firstly produced according the key points (namely head locations, of which the positions of key points are $1$, other pixels' values are $0$). Then we adopt a ball-shaped structuring element with radius and height of $50$ pixels to dilate the binary image, which is an image morphology operation. Finally, we obtain a coarse segmentation mask.

Since the DME and FBS are pixel-wise tasks, they share a base network as the feature extractor. In Fig. \ref{Fig-overview}, the orange feature maps represent the data flow of the segmentation stream. The last feature map is appended to the last feature map in DME to estimate the density map. Unfortunately, the outputs of two tasks are not in the same order of magnitude, which caused that the feature map from the two streams are also not in the same order. Thus, it is difficult to train the model only by the direct concatenation of two feature maps. To this end, we design a transition layer to tackle this problem. As shown in the Fig. \ref{Fig-overview}, two transition layers are added to the beginning and end of segmentation stream. As a matter of fact, the transition layer is a standard convolutional layer with $1 \times 1$ kernel size, and it produces the feature map with the same size as the input. 

At the training stage, the loss function of FBS is the standard 2-D cross-entropy loss.

\subsection{Perspective Encoder: DULR Module}
\label{Sec_DULR}
After introducing FBS, the model can extract the three-level features, namely the global feature of R-HDC, the middle-level feature from FBS and the local feature extracted by DME. However, the model cannot effectively handle the perspective changes problem mentioned in Section \ref{Sec_intro}. To this end, we exploit the Spatial CNN \cite{pan2017spatial} to encode the perspective changes on four directions, namely Down, Up, towards the Left and Right, which is also called as ``DULR module''.

\begin{figure*}
	\centering
	\includegraphics[width=0.85\textwidth]{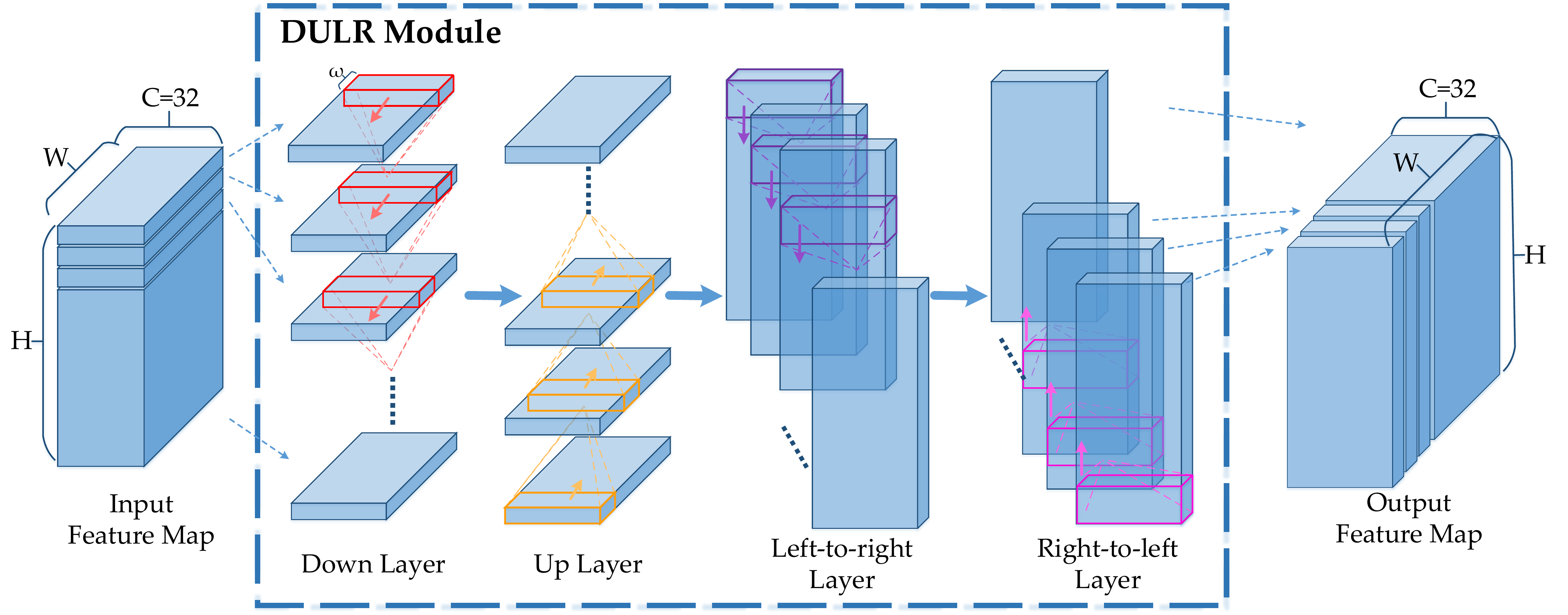}
	\caption{The detailed architecture of DULR module.}\label{Fig-DULR}
\end{figure*}

Fig. \ref{Fig-DULR} illustrates the detailed architecture of DULR module, which consists of four convolutional layers (Down, up, Left-to-right and Right-to-left layers) to respectively handle four directions. 
As shown in Fig. \ref{Fig-DULR}, taking the Down Layer as an example, it consists of a convolutional operation with C kernels of size $C \times \omega $ and a ReLU activation function. The concrete operation will be demonstrated below. Firstly, the feature map $F$ with size of $C \times H \times W$ is divided into $H$ parts with the size of $C \times W$, which are represented by $F_H^i$ ($i \in [1,H]$), where $C$, $H$ and $W$ are respectively the size of channel, height and width, $i$ denotes the index of the $H$ parts. Next, the Down Layer is applied on $F_H^1$ and produces a $(F\_D)_H^i$ with the same size as $F_H^1$. Then the sum of $(F\_D)_H^i$ and the next part $F_H^2$ is fed into the Down Layer to obtain $(F\_D)_H^2$. Iteratively, the $H$ parts of $(F\_D)_H^i$ will be output. Finally, they are concatenated to become a feature map $F\_D$ with the original size of $F$ ($C \times H \times W$).

For showing the operation of Down Layer intuitively, the formulated feed-forward computation is defined as below:

\begin{equation}
\begin{array}{l}
\begin{aligned}
(F\_D)_H^i = \left\{ \begin{array}{l}
\emph{down}(F_H^i),\quad i = 1\\
\emph{down}(F_H^i + \emph{down}(F_H^{i - 1})),\quad i = 2,3,...,H
\end{array} \right.,
\end{aligned}\label{PDC}
\end{array}
\end{equation}
where $\emph{down}(*)$ denotes the operation Down Layer (Conv + ReLU). It is noted that the $H$ parts of $F_H^i$ share the same Down Layer, which is also treated as a kind of  recurrent neural network. Similarly, the other three layers (Up, Left-to-right and Right-to-left Layer) have the similar operation except for the sliding direction. 

As for a crowd scene, given a row, its result considers the results of the rows above it. In fact, it is an aggregation of density feature representation. As for different rows, because of different computing orders, the effects of aggregation for each row are different, which is a one-to-one correspondence with the change in perspective. Finally, through DULR, the feature map potentially contains the perspective changes of the entire image. In fact, the perspective changes are treated as the global contextual information. According to the mechanism of DULR, the representation of a specific region contains the global information with different extent. Thus, the representation is able to more accurately predict the density map.

In summary, the DULR module remains the original size of the input, which introduces the global spatial information into the whole feature map. The operation effectively encodes the perspective changes of crowd scenes, especially extremely congested regions. 

\section{Experiments}

In this section, we firstly describe the evaluation metrics and the experimental details. Then, the ablation studies are applied on the ShanghaiTech Part A dataset. Finally, we report the results of the proposed PCC Net on the four mainstream datasets.

\subsection{Evaluation}

In the field of crowd counting, the mainstream evaluation metrics are  Mean Absolute Error (MAE) and Mean Squared Error (MSE), which are defined as below:

\begin{equation}
\begin{array}{l}
\begin{aligned}
MAE = \frac{1}{N}\sum\limits_{i = 1}^N {\left| {{y_i} - {{\hat y}_i}} \right|},
\end{aligned}\label{MAE}
\end{array}
\end{equation}

\begin{equation}
\begin{array}{l}
\begin{aligned}
MSE = \sqrt {\frac{1}{N}\sum\limits_{i = 1}^N {{{\left| {{y_i} - {{\hat y}_i}} \right|}^2}} } ,
\end{aligned}\label{MSE}
\end{array}
\end{equation} 
where $N$ is the number of samples in test set, ${{y_i}}$ is the count label and ${{{\hat y}_i}}$ is the estimated count value for the $i$th test sample. In addition, we also evaluate the quality of density maps using PSNR (Peak Signal-to-Noise Ratio) ans SSIM (Structural Similarity in Image \cite{wang2004image}) in Section \ref{comSOA}.

\subsection{Experimental setup}

\subsubsection{Implementation Details} The full model (PCC Net) aims to optimizing the loss function as follows:
\begin{equation}
\begin{array}{l}
\begin{aligned}
\mathcal{L} = \mathcal{L}_{DME} + \lambda \mathcal{L}_{R-HDC} + \beta \mathcal{L}_{FBS}
\end{aligned}\label{loss}.
\end{array}
\end{equation}
where $\mathcal{L}_{DME}$, $\mathcal{L}_{R-HDC}$ and $\mathcal{L}_{FBS}$ denote the loss function of DME, R-HDC and FBS, respectively. The concrete definitions are mentioned in Section \ref{Sec_DME}, \ref{Sec_R-HDC} and \ref{Sec_FBS}. During the training stage, the $\lambda$ and $\beta$ are set as ${10^{ - 4}}$. The learning rates of DME and R-HDC is initialized at ${10^{ - 4}}$, and FBS's learning rate is set as ${10^{ - 2}}$. Note that all learning rates are fixed. The batch is $12$. In the R-HDC, the number of ROIs is $20$. And in the DULR module, the kernel width $ \omega $ is set as $1$. All images are resized to $576 \times 768$, and the labels are generated under the same size. Finally, we adopt Adam algorithm to optimize PCC Net and obtain the best results. 

The training and evaluation are performed on NVIDIA GTX 1080Ti GPU using PyTorch framework \cite{paszke2017pytorch}.

\subsubsection{Data augmentation}

Original training images (size of $576 \times 768$) are randomly cropped to the size of $512 \times 680$. Accordingly, density maps and segmentation maps are correspondingly cropped, and the count labels are recalculated. In addition, random horizontally flipping is employed in the training stage.

\subsection{Ablation Experiments on ShanghaiTech Part A}
\label{ablation}

\begin{figure*}[htbp]
	\centering
	\includegraphics[width=0.95\textwidth]{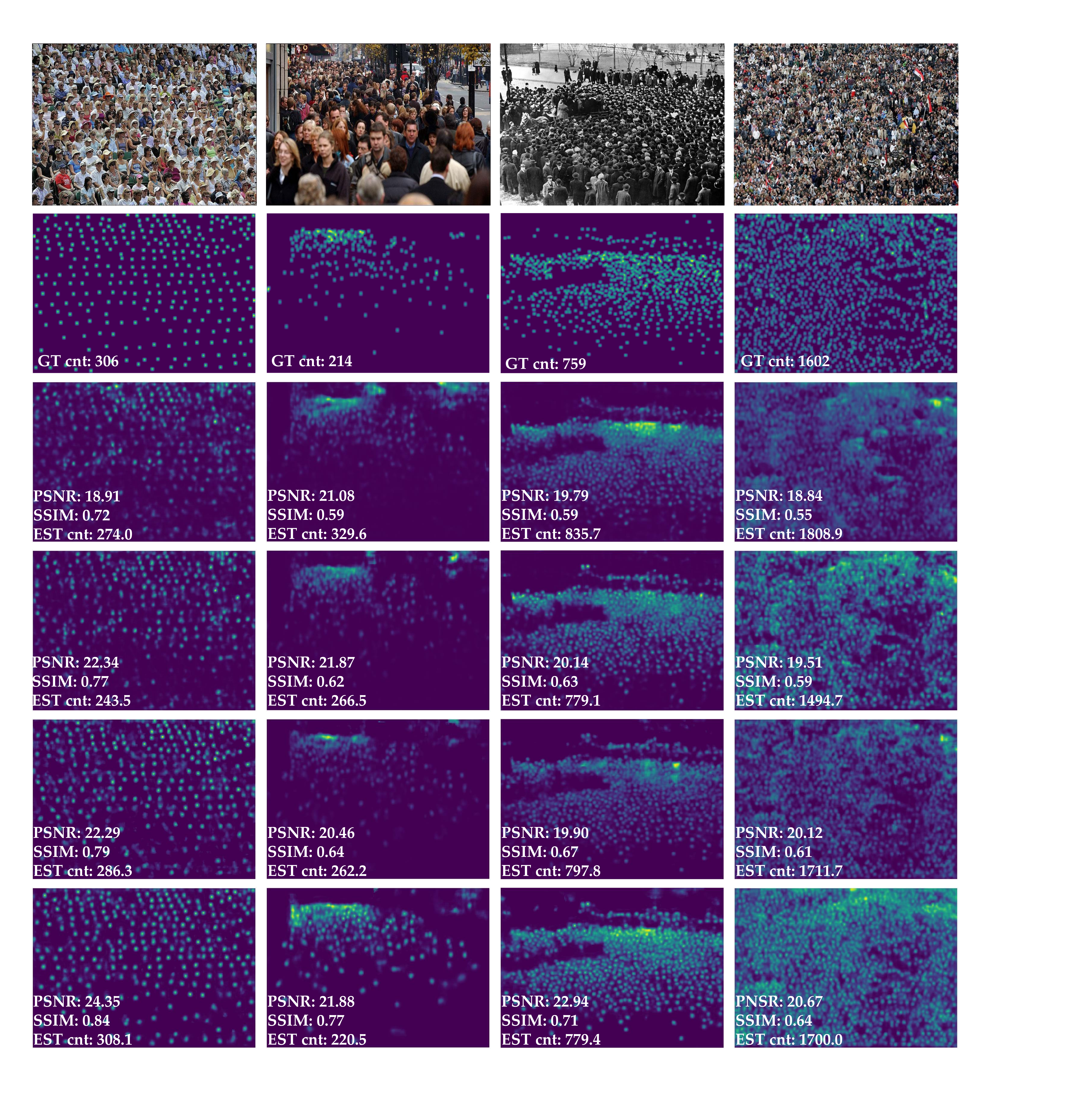}
	\caption{Exemplar results of step-wise models on Shanghai Tech Part A dataset. The first row shows the original images. The heat map in the second row is ground truth. The last four rows demonstrate the results of DME, DME+R-HDC, DME+R-HDC+FBS and DME+R-HDC+FBS+DULR (PCC Net). ``GT cnt'' and ``EST cnt'' denote the ground truth count and the estimation count, respectively.}\label{Fig_show}
\end{figure*}

In order to demonstrate the effects of the proposed method, the ablation experiments are applied on ShanghaiTech Part A Dataset \cite{zhang2016single}, which is from ShanghaiTech Dataset. It contains 300 training images and 182 testing images with different sizes. The images are collected from the Internet. The concrete ablation study as below: the step-wise experiments (DME, DME+R-HDC, DME+R-HDC+FBS and DME+R-HDC+FBS+DULR) are firstly reported. Furthermore, the comparative experiments are conducted to analyze the effect of the proposed R-HDC and DULR Module. 

\subsubsection{The Results of Step-wise Experiments}

Table \ref{Table-ablation} illustrates the MAE and MSE of the proposed method. To be specific, 1) DME: only single task to predict the density map; 2) DME+R-HDC: a multi task with R-HDC; 3) DME+R-HDC+FBS: based on 2), the FBS is considered as another training task; 4) DME+R-HDC+FBS+DULR (PCC Net): the DULR module is added to the architectures of DME and FBS.  From Table \ref{Table-ablation}, each step strategy significantly improves the performance of the model. The proposed full model, PCC Net, achieves the best of MAE (73.5) and MSE (124.0). 

\begin{table}[htbp]
	
	\centering
	\caption{Estimation errors for step-wise models of the proposed method on ShanghaiTech Part A.}
	
	\begin{tabular}{c|cc}
		\whline
		Methods 				 &MAE 		&MSE	 		\\
		\whline
		DME           &106.9 &169.3	\\
		\hline
		DME+HDC     &95.6   &147.6 	\\
		DME+R-HDC     &88.9   &137.2 	\\
		\hline
		DME+R-HDC+FBS   &80.1 &124.4  \\
		\hline
		DME+R-HDC+FBS+4conv   &79.4	& 130.6
		\\
		DME+R-HDC+FBS+DULR (PCC Net)   &\textbf{73.5} &\textbf{124.0}  \\
		\whline
		
	\end{tabular}\label{Table-ablation}
\end{table}

Fig. \ref{Fig_show} presents some visualization and crowd counting results. The first row shows the input images. The heat map in the second row is ground truth. And the last four rows demonstrate the results of of the above four methods. ``GT cnt'' and ``EST cnt'' in density maps denote the ground truth count and the estimation count, respectively. The input image in Column 1 has slight perspective change. The first three results show the inconsistent density for different regions. However, the results of PCC Net show the closely uniform density map, which means that introducing the DULR module effectively extracts the uniform perspective feature. From Column 2, Row 5, FBS can reduce the background errors. Column 3 and 4 show the results of the two congested crowd scenes. From them, the intuitive improvements of each step are demonstrated.

\subsubsection{The Effect of R-HDC}
In this section, we compare the estimation errors of HDC \cite{sindagi2017cnn} and the proposed random HDC (R-HDC). To be specific, the comparing experiment (DME+HDC) is designed as below. According the experimental configuration of \cite{sindagi2017cnn}: the fixed training patches are generated for the HDC, and adopting Spatial Pyramid Pooling (SPP) \cite{he2014spatial} to adapt feature maps to a specific size. Note that the network architecture of DME+HDC is the same as that of DME+R-HDC except for the above differences. The other training details (such as learning rate, optimizer and so on) are keep with \cite{sindagi2017cnn}.

The second box in Table \ref{Table-ablation} reports the results of DME+HDC and DME+R-HDC. The two estimation errors of the latter (MAE: 95.6, MSE: 147.6) are less than that of the former (MAE: 88.9, MSE: 137.2). By iteratively training phase, R-HDC fully covers the whole training samples at the high level. Besides, R-HDC randomly crops the feature maps to patches with a random size, so it can be regarded as data augmentation. Thus, for the high-level feature extraction, R-HDC's learning ability significantly outperforms the traditional HDC.

\subsubsection{The Effect of DULR Module} 
In order to demonstrate the effect of the DULR module, the extensive experiments are conducted in this section. To be specific, the DULR module is replaced by four sequential standard convolutional layers, which is called as ``4conv''. The 4conv's kernel size is the same as that of the DULR module. In other word, the slice-by-slice convolutional operations on the four directions are removed from DULR module. Thus, the 4conv cannot encode the spatial information to describe the perspective changes in the crowd scenes.

The last box in Table \ref{Table-ablation} reports the results of DME+R-HDC+FBS+4conv and DME+R-HDC+FBS+DULR (PCC Net). Compared with DME+R-HDC+ FBS, adding 4conv to it does not significantly reduce the estimation errors (MAE: from 80.1 to 79.4). On the contrary, the MSE slightly increases from 124.4 to 130.6. As for the DULR module, the estimation errors reduce remarkably, whether MAE (from 80.1 to 73.5) or MSE (from 124.4 to 124.0). This experimental result verifies the effectiveness of the DULR module and shows the importance of encoding perspective changes.

\subsection{Results on ShanghaiTech}

ShanghaiTech dataset is a large-scale crowd counting dataset, which consist of part A and B. The detailed description of Part A is shown in \ref{ablation}. Part B is collected from the surveillance camera in metropolitan areas. It contains 400 training images and 316 test images with the same resolution of $768 \times 1024$.

Table \mbox{\ref{shanghai}} lists the results of some mainstream methods (our PCC Net, Zhang \mbox{\emph{et al.}} \mbox{\cite{zhang2015cross}}, MCNN \mbox{\cite{zhang2016single}}, FCN \mbox{\cite{zhang2016single}}, Cascaded-MTL \mbox{\cite{sindagi2017cnn}}, Switching-CNN \mbox{\cite{sam2017switching}}, CP-CNN \mbox{\cite{sindagi2017generating}}, ACSCP \mbox{\cite{shen2018crowd}} and so on) on ShanghaiTech Part A and B datasets. Compared with other no-pre-trained methods, PCC Net achieves the best MAE of 73.5 and the second-best MSE of 124.0 on Part A. On Part B, our method outperforms the state-of-the-art method (ACSCP \mbox{\cite{shen2018crowd}}) and achieves an amazing result: MAE of 11.0 (6.2-point improvement) and MSE of 19.0 (8.4-point improvement). From the all list, our PCC Net attains a competitive result (forth/second place) on Part A/B, respectively.

\begin{table}[htbp]
	
	\centering
	\caption{Estimation errors on ShanghaiTech dataset.}
	\setlength{\tabcolsep}{1.60mm}{
	\begin{tabular}{c|c|cc|cc}
		\whline
		\multirow{2}{*}{Method}&\multirow{2}{*}{PrTr} &	\multicolumn{2}{c|}{Part A} 	&		\multicolumn{2}{c}{Part B} 	\\
		\cline{3-6} 
		
			&			 &MAE 		&MSE			&MAE			&MSE 		 		\\
		\whline
		Zhang \emph{et al.} \cite{zhang2015cross}	&\ding{55} &181.8 	&277.7	&32.0 &49.8			\\
		\hline
		MCNN \cite{zhang2016single}	&\ding{55}&110.2& 173.2& 26.4 &41.3 \\
		\hline
		FCN \cite{marsden2016fully}	&\ding{55} &126.5& 173.5& 23.8 &33.1 \\
		\hline
		Cascaded-MTL \cite{sindagi2017cnn}	&\ding{55} &101.3 &152.4 &20.0 &31.1 \\	
		\hline
		ACSCP \cite{shen2018crowd}	&\ding{55} &75.7  &\textbf{\underline{102.7}}  &17.2 & 27.4  \\
		\hline	
		DecideNet \cite{liu2018decidenet}	&\ding{55} &-  &-  &20.75 & 29.42  \\
		\hline
		\textbf{PCC Net (ours)} 	&\ding{55} &\underline{73.5} 	&124.0 &\underline{11.0}	&\underline{19.0}  \\
		\whline
		Switching-CNN \cite{sam2017switching}	&\ding{51} &90.4& 135.0& 21.6 &33.4 \\
		\hline
		CP-CNN \cite{sindagi2017generating}	&\ding{51} &73.6  &\underline{106.4}  &20.1 & 30.1  \\
		\hline
		CSRNet \cite{li2018csrnet}	&\ding{51} &\textbf{\underline{68.2}}  &115.0  &\textbf{\underline{10.6}} & \textbf{\underline{16.0}}  \\
		\hline
		IG-CNN \cite{babu2018divide}	&\ding{51} &72.5  &118.2  &13.6 & 21.1  \\
		\hline
		D-ConvNet \cite{shi2018crowd}	&\ding{51}  &73.5  &112.3  &18.7  &26.0  \\
		\hline
		L2R \cite{liu2018leveraging}	&\ding{51}  &72.0  &106.6  &14.4 & 23.8  \\
		\whline
		
	\end{tabular}
	}
\label{shanghai}
\end{table}


\subsection{Results on WorldExpo'10}

WorldExpo'10 dataset is presented by Zhang \emph{et al.} \cite{zhang2015cross}, which is a cross-scene large-sale dataset. The data are taken from 108 surveillance cameras in Shanghai 2010 WorldExpo event, containing $3,980$ images with size of $576 \times 720$ and 199,923 labeled pedestrians. The single scene contains the no more than 220 pedestrians, so it is not extremely dense crowds scenes. The training set is from 103 scenes totaling $3,380$ images. And the test set is from the rest 5 scenes. Each test scene have 120 images. In addition, the dataset provides the perspective map and ROI map. 

\begin{table}[htbp]
	
	\centering
	\caption{Estimation errors on WorldExpo'10.}
	\setlength{\tabcolsep}{1.00mm}{
	\begin{tabular}{c|c|ccccc|c}
		\whline
		Methods 		&PrTr			 &S1 &S2 &S3 &S4 &S5 &Avg. 	 		\\
		\whline
		Chen \emph{et al.} \cite{chen2013cumulative}&\ding{55} &2.1& 55.9 &9.6 &11.3 &3.4 &16.5		\\
		\hline
		Zhang \emph{et al.} \cite{zhang2015cross}&\ding{55} &9.8 &14.1 &14.3 &22.2 &3.7 &12.9	\\
		\hline
		MCNN \cite{zhang2016single}&\ding{55}&3.4 &20.6 &12.9 &13.0 &8.1 &11.6 \\
		\hline
		Cascaded-MTL \cite{sindagi2017cnn}&\ding{55} &3.8  &32.3  &19.5  &20.6  &6.6  &16.6 \\
		\hline
		ACSCP \cite{shen2018crowd}	&\ding{55} & 2.8&	14.05&	9.6	&\textbf{\underline{8.1}}&	\underline{2.9}	&\textbf{\underline{7.5}}  \\
		\hline	
		DecideNet \cite{liu2018decidenet}	&\ding{55}& 2.0	&\underline{13.14}	&\underline{8.9}	&17.4	&4.75	&9.23 \\
		\hline
		\textbf{PCC Net (ours)} &\ding{55} &\textbf{\underline{1.9}} &18.3  &10.5 &13.4 &3.4 &9.5 \\
		\whline
		Shang \emph{et al.} \cite{shang2016end} 	&\ding{51} &7.8 &15.4 &14.9 &11.8 &5.8 &11.7 \\
		\hline
		Switching-CNN \cite{sam2017switching}	&\ding{51} &4.4 &15.7 &10.0 &11.0 &5.9 &9.4 \\
		\hline
		CP-CNN \cite{sindagi2017generating}	&\ding{51} &2.9 &14.7 &10.5 &10.4 &5.8 &8.9   \\
		\hline
		CSRNet \cite{li2018csrnet}	&\ding{51}& 2.9&	\textbf{\underline{11.5}}&	\textbf{\underline{8.6}}&	16.6&	3.4&	\underline{8.6}  \\
		\hline
		IG-CNN \cite{babu2018divide}	&\ding{51}& 2.6&	16.1&	10.15&	20.2&	7.6&	11.3  \\
		\hline
		D-ConvNet \cite{shi2018crowd}	&\ding{51} & \textbf{\underline{1.9}}&	12.1&	20.7&	\underline{8.3}&	\textbf{\underline{2.6}}&	9.1  \\
		\whline
		
	\end{tabular}
	}
	
	\label{WE}
\end{table}

Table \mbox{\ref{WE}} reports the Mean Absolute Errors (MAE) of the proposed PCC Net and some state-of-the-art algorithms. To be specific, the MAE of five scenes and their average are listed in the table. From it, the MAE of Scene1 (1.9) is the best. As for the average, the proposed PCC Net is the third prize in the no-pre-trained methods, which is close to the top-2 best methods. The phenomenon is different from the results on Shanghai Tech and  UCF\_CC\_50 dataset. The main reasons are: 1) the effect of the DULR module for low density scenes is limited; 2) the dataset provides the perspective map so that all methods can directly encode the perspective changes.


\subsection{Results on UCF\_CC\_50}

UCF\_CC\_50 dataset is developed by Idrees \emph{et al.} \cite{idrees2013multi} from University of Central Florida, which only contains 50 images but has 63,075 labeled individuals. It includes a wide range of densities (the range of individuals from 94 to 4,543) and cover diverse scenes with varying perspective distortion. Because of only 50 images, we employ the standard 5-fold cross-validation protocol to evaluate the algorithm, which is also adopted by other state-of-the-art methods.

Table \mbox{\ref{UCF}} lists the estimation errors of the proposed PCC Net and some state-of-the-art algorithms. The proposed PCC Net significantly outperforms the other methods. Especially, PCC Net reports a 26.1-point improvement in MAE and a 82.0-point improvement in MSE over CSRNet \mbox{\cite{li2018csrnet}}. The significant improvements evidence the proposed PCC Net can effectively handle extremely congested crowd scenes.

\begin{table}[htbp]
	
	\centering
	\caption{Estimation errors on UCF\_CC\_50 dataset.}
	
	\begin{tabular}{c|c|cc}
		\whline
		Methods 	&PrTr			 &MAE 		&MSE	 		\\
		\whline
		Idrees \emph{et al.} \cite{idrees2013multi}&\ding{55} &419.5& 541.6		\\
		\hline
		Zhang \emph{et al.} \cite{zhang2015cross}&\ding{55} &467.0& 498.5	\\
		\hline
		MCNN \cite{zhang2016single}&\ding{55}&377.6& 509.1 \\
		\hline
		FCN \cite{marsden2016fully}&\ding{55}&338.6& 424.5 \\
		\hline
		Onoro \emph{et al.} \cite{onoro2016towards} Hydra-2s &\ding{55} &333.7 &425.2 \\
		\hline
		Onoro \emph{et al.} \cite{onoro2016towards} Hydra-3s&\ding{55}&465.7 &371.8 \\
		\hline				
		Walach \emph{et al.} \cite{walach2016learning}&\ding{55}&364.4& 341.4 \\
		\hline			
		Cascaded-MTL \cite{sindagi2017cnn}&\ding{55} &322.8& 341.4 \\
		\hline
		ACSCP \cite{shen2018crowd}&\ding{55} &291.0& 404.6 \\
		\hline		
		\textbf{PCC Net (ours)}&\ding{55}  &\textbf{\underline{240.0}}	&\textbf{\underline{315.5}} \\
		\whline
		Switching-CNN \cite{sam2017switching}&\ding{51} &318.1 &439.2 \\
		\hline
		CP-CNN \cite{sindagi2017generating}&\ding{51} &295.8& \underline{320.9}   \\
		\hline		
		CSRNet \cite{li2018csrnet}&\ding{51} &\underline{266.1}	&397.5   \\
		\hline
		IG-CNN \cite{babu2018divide}&\ding{51} &291.4&	349.4  \\
		\hline
		D-ConvNet \cite{shi2018crowd}&\ding{51} &288.4&	404.7   \\
		\hline
		L2R \cite{liu2018leveraging}&\ding{51} &279.6&	388.9   \\		
		\whline
		
	\end{tabular}\label{UCF}
\end{table}

\subsection{Results on UCF-QNRF}

UCF-QNRF \mbox{\cite{idrees2018composition}} is a large-scale extremely congested crowd counting dataset, which consists of 1,535 crowd images captured from 
Flickr, Web Search and the Hajj footage. It is the most large-scale crowded dataset, of which the count number is in range from 49 to 12,865. From the Internet, \mbox{\cite{idrees2018composition}} manually searches the following keys: CROWD, HAJJ, SPECTATOR CROWD, PILGRIMAGE, PROTEST CROWD and CONCERT CROWD. Through the above keys, the crowd scenes can be effectively collected. 

Table \mbox{\ref{UCF-QNRF}} reports the performance of PCC Net and other mainstream algorithms. Compared with the three no-pre-trained methods (Idrees \mbox{\emph{et al.}} \mbox{\cite{idrees2013multi}}, MCNN \mbox{\cite{zhang2016single}} and Cascaded-MTL \mbox{\cite{sindagi2017cnn}}), we achieve the best result (MAE of 148.7 and MSE of 247.3). The PCC Net even outperforms Switching-CNN \mbox{\cite{sam2017switching}}, a method based on VGG-16 pretrained model. In the list, CL \mbox{\cite{idrees2018composition}} attains the lowest estimation errors (MAE of 132 and MSE of 191). Considering that CL adopts DenseNet-201, our PCC Net is competitive in some way. 

\begin{table}[htbp]
	
	\centering
	\caption{Estimation errors on UCF-QNRF dataset.}
	
	\begin{tabular}{c|c|cc}
		\whline
		Methods 	&PrTr		 &MAE 		&MSE	 		\\
		\whline
		Idrees \emph{et al.} \cite{idrees2013multi}&\ding{55}  & 315&  508		\\
		\hline
		MCNN \cite{zhang2016single}&\ding{55}&277&426  \\
		\hline
		Cascaded-MTL \cite{sindagi2017cnn}&\ding{55} &252& 514 \\
		\hline
		\textbf{PCC Net (ours)} &\ding{55} &\underline{148.7}	& \underline{247.3} \\
		\whline
		Switching-CNN \cite{sam2017switching}&\ding{51} &228 &445 \\
		\hline
		CL\cite{idrees2018composition} &\ding{51} &\textbf{\underline{132}}	& \textbf{\underline{191}} \\
		\whline
	\end{tabular}\label{UCF-QNRF}
\end{table}

\section{Discussion and Analysis}

\subsection{Comparisons with the State-of-the-art Methods}

\label{comSOA}

\begin{table*}[htbp]
	\centering
	\caption{The detailed information of the proposed model and the state-of-the-art methods.}	
	\begin{tabular}{c|cc|cc|ccc|c}
		\whline
		Methods 	&MAE 	&MSE &PSNR &SSIM &Model Size & Params &	Runtime (ms) & pre-train		\\
		\whline
		Cascaded-MTL \cite{sindagi2017cnn} &126.5	&173.5 &- &- &0.5M &0.12M &3 &\ding{55}\\				
		\hline
		Switching-CNN \cite{sam2017switching}  &90.4 &135.0 &21.91 &0.67 &57.6MB &15.1M &153 &\ding{51}\\
		\hline
		CP-CNN \cite{sindagi2017generating} &73.6 & \textbf{106.4}  &21.72 &0.72 &$>$500MB &62.9M &5113&\ding{51}\\
		\hline
		PCC Net (ours)  &\textbf{73.5}	&124.0 &\textbf{22.78} &\textbf{0.74} &2.0MB &0.55M &89 &\ding{55}\\
		\whline		
	\end{tabular}
	\label{comparison}
\end{table*}

In order to show the superiority of our methods, we compared the details of the proposed model with the state-of-the-art methods (Cascaded-MTL \mbox{\cite{sindagi2017cnn}}, Switching-CNN \cite{sam2017switching} and CP-CNN \cite{sindagi2017generating}). Table \ref{comparison} lists the four main metrics to evaluate density estimation performance: MAE, MSE, PSNR (Peak Signal-to-Noise Ratio) ans SSIM (Structural Similarity in Image \cite{wang2004image}) on the ShanghaiTach Part A. From it, PCC Net is the best except for MSE.

In addition, the computation performance (model size, number of parameters and runtime) is also shown in the table. Although Cascaded-MTL \mbox{\cite{sindagi2017cnn}} is the lightest model in \mbox{\cite{sam2017switching, sindagi2017generating}} and PCC Net, its estimation result is the poorest in them. The performance of these three algorithms is roughly at the same level. Thus, we focus on comparing these three methods. In general, PCC Net is the best on these three metrics: its model size and number of parameters is respectively only 2.0MB and 0.55M, which are far less than that of Switching-CNN and CP-CNN. As for the runtime, PCC Net is also faster than that of them (89ms versus 153ms and 5113ms). Note that the test is implemented on the one NVIDIA GTX 1080Ti GPU. Furthermore, the PCC Net is trained from scratch, which does not need the pre-trained model. However, Switching-CNN and CP-CNN adopt the pre-trained model on ImageNet. 

In general, considering the performance and model size, the proposed PCC Net is very competitive.

\subsection{Analysis of the Loss Weights}

\begin{figure}[htbp]
	\centering
	\begin{subfigure}{0.23\textwidth} 
		\includegraphics[width=\textwidth]{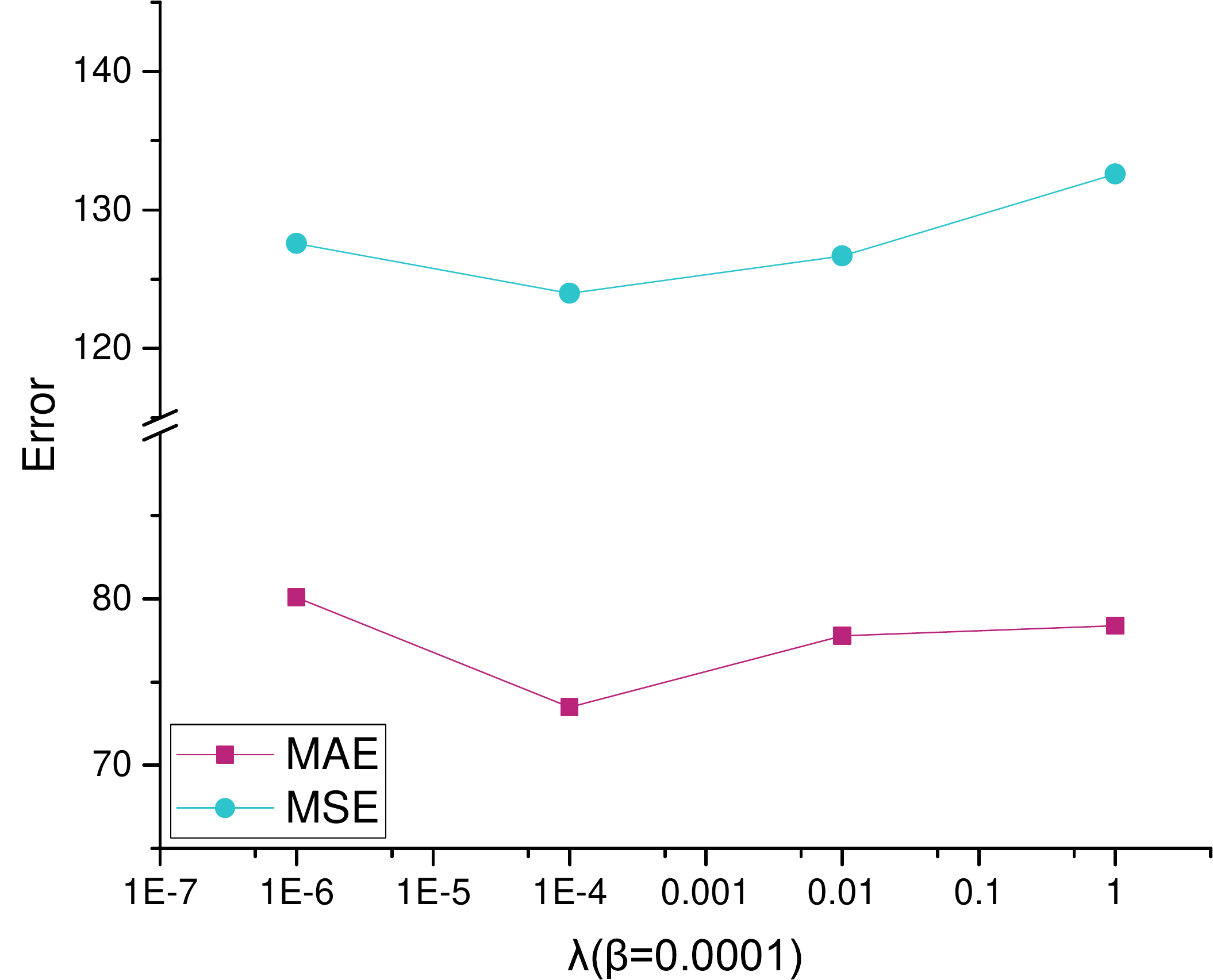}
		\caption{line chart of performance under different lamda.} 
	\end{subfigure}
	\vspace{1em} 
	\begin{subfigure}{0.23\textwidth} 
		\includegraphics[width=\textwidth]{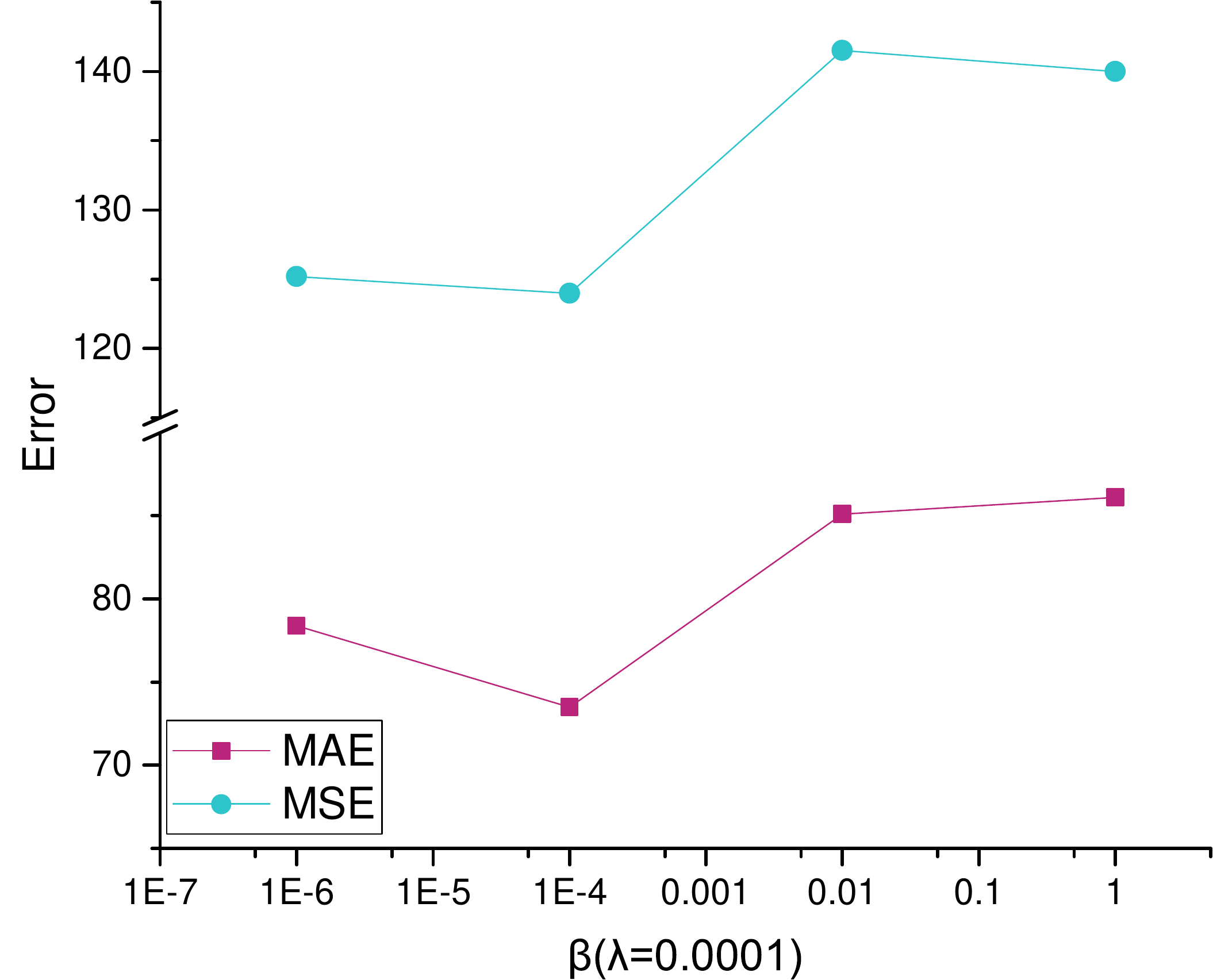}
		\caption{line chart of performance under different beta.} 
	\end{subfigure}
	\caption{The performance under the different loss weights in Eq. \ref{loss}.} \label{lamda}
\end{figure}

In this section, we discuss that how the multi-loss weights affect the estimation performance. To be specific, we set the $\lambda$ and $\beta$ in $\left\{ {{\rm{1,1}}{{\rm{0}}^{{\rm{ - 2}}}}{\rm{,1}}{{\rm{0}}^{{\rm{ - 4}}}}{\rm{,1}}{{\rm{0}}^{{\rm{ - 6}}}}} \right\}$ to train PCC Net. Fig. \ref{lamda} shows the performance (MAE and MSE) under the different values. From it, the model achieves the best MAE and MSE when $\lambda$ and $\beta$ are set as $0.0001$. What's more, we find the $\beta$ affects the density estimation more dramatically than $\lambda$. The main reason is FBS and DME share more conv layers than those between R-HDC and DME. Furthermore, DME is a regression problem, which is more difficult than R-HDC and FBS. Thus, the weights of DME is larger than others.

\subsection{Discussions of the DULR Module}

In order to analyze each module and different combinations, we conduct eight groups of comparative experiments, of which the results are shown in Table \mbox{\ref{comDULR}}. Because of different camera angles and parameters, vanishing points (VP) locate at different positions in crowd scenes. Generally, the regions that are closer to VP have larger density. Thus, 4-direction DULR provides more perspective information than single DU and the single module. At the same time, we also find that the results of using the same number of modules are very close.

\begin{table}[htbp]
	
	\centering
	\caption{Performance of each module in DULR on ShanghaiTech Part A Dataset.}
	\setlength{\tabcolsep}{0.90mm}{
	\begin{tabular}{c|c|c|c|c|c|c|c|c}
		\whline
		 	&4conv 	&D	&U &L &R &DU &LR & DULR	\\
		\whline
		MAE&79.4 &76.9 &77.3 &77.6 &77.1 &74.7 &75.1 &\textbf{73.5} \\
		\hline
		MSE&130.6 &129.2 &126.5 &130.4 &127.4 &126.4 &126.9 &\textbf{124.0} 	\\
		\whline	
	\end{tabular}
	}
	\label{comDULR}
\end{table}

For directly explaining the DULR's effect for encoding perspective map, the experiment on WorldExpo'10 is conducted (only WorldExpo provides the perspective map). To be specific, we compare two CNNs (DME+4conv and DME+DULR, both have the same number of parameters), which directly output the perspective map according to the input image. Table \ref{pMap} shows the performance (MAE and MSE) of two CNNs. From it, we find the DME+DULR predicts better than DME+4conv, which directly verify that DULR can more effectively encode the perspective changes than traditional convolutional operation.

\begin{table}[htbp]
	
	\centering
	\caption{Results of perspective map estimation errors.}
	
	\begin{tabular}{c|cc}
		\whline
		Methods &DME+4conv 	&DME+DULR		\\
		\whline
		MAE &16.82& \textbf{10.36}		\\
		\hline
		MSE  &23.45& \textbf{14.48}	\\
		\whline
		
	\end{tabular}\label{pMap}
\end{table}

\begin{figure}
	\centering
	\includegraphics[width=0.48\textwidth]{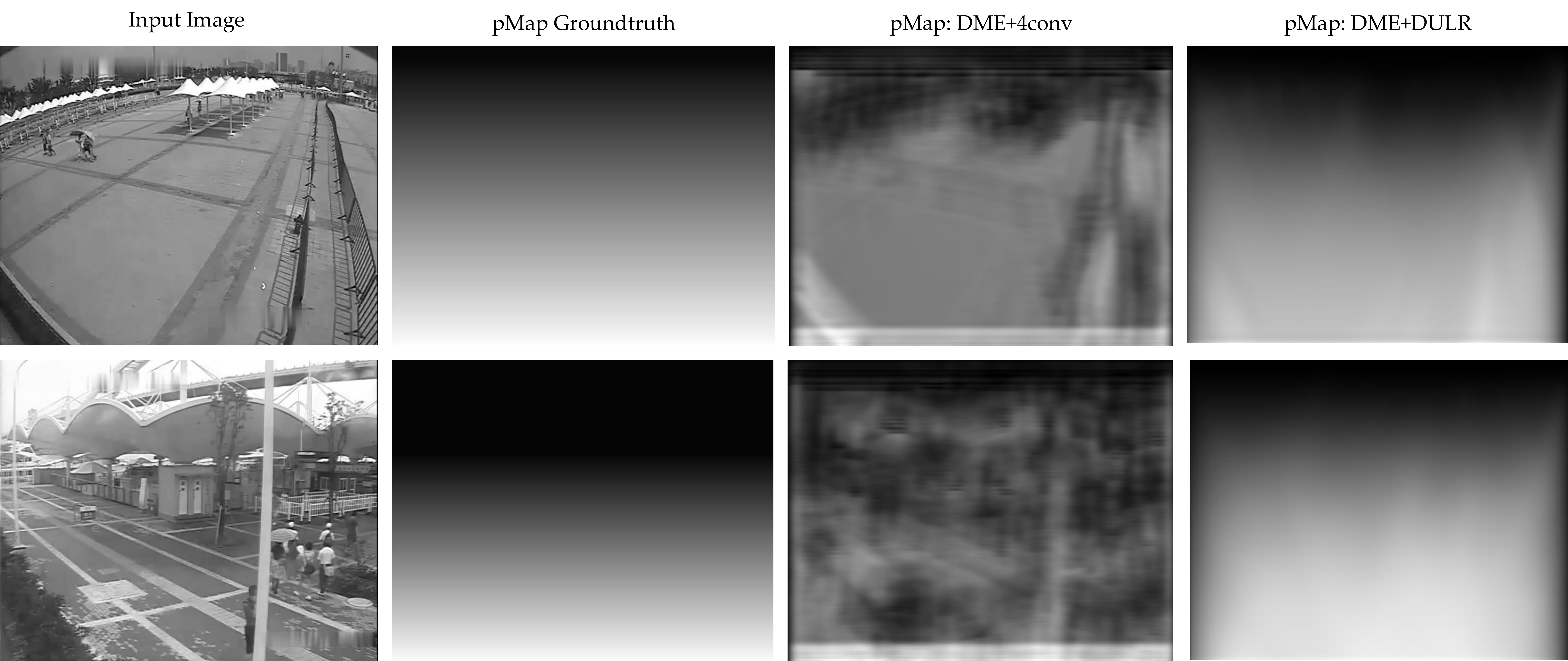}
	\caption{Exemplar results of perspective map (pMap) estimation on WorldExpo'10 dataset.}\label{Fig_pmap}
\end{figure}

Fig. \mbox{\ref{Fig_pmap}} shows the visualization results of the perspective map (pMap). Intuitively, DME+DULR produces the more high-quality pMap than DME+4conv. The latter's outputs can not encode the global information and perspective changes, which causes it is sensitive to local features. As for the results of DME+DULR, it can effectively show the gradual changes of perspective in the scenes.

\subsection{Selection of the ROI's Number}

In R-HDC, an important parameter is the number of patches for ROI Pooling \mbox{\cite{7410526}}. In Faster RCNN \mbox{\cite{7410526}}, the authors generate 300 ROIs by a Region Proposal Network (RPN), which is a trainable Fully Convolutional Network (FCN). Since Faster RCNN belongs to a task of object detection, the purpose of extracting 300 ROIs is to cover as many as possible region candidates to avoid missing any potential object. Different from it, R-HDC aims to randomly generate patches and pool them to a consistent size by ROI Pooling, which are not from a learnable network. In other words, the generated ROIs by R-HDC do not have a real object category. Thus, we only try to ensure that the selected patches can cover the whole image. By iterative training (~800 epochs), for each image, R-HDC also generates 1,600 different patches in theory. Thus, we do not need so many patches like Faster RCNN \mbox{\cite{7410526}}.

Here, we discuss that the selection of the parameter. In the experiments, the area of each generated patch is more than 1/16 that of the input image, which is explained in Section \mbox{\ref{Sec_R-HDC}}. In an extreme situation, all patches are 1/16 size of the original image. Then we need at least 16 no-overlapping patches to cover the full image. However, no matter how many image blocks are generated with random locations, it is impossible to 100\% guarantee that they can cover each entire image in theory. In order to select an optimal parameter, we conduct the experiments under different numbers of ROIs. To be specific, we conduct a group of experiments on Shanghai Tech B under different number of ROIs, namely 5, 10, 20 (our final selection), 30, 40, 50, 100, 200 and 300 (Faster RCNN’s selection). From the results in Table \mbox{\ref{ROIs}}, we find the performance does not increase obviously when the number is more than 20. Thus, considering the training speed and GPU memory, we finally set the number as 20.

\begin{table*}[htbp]
	
	\centering
	\caption{Performance under different number of ROIs on Shanghai Tech B dataset.}
	
	\begin{tabular}{c|c|c|c|c|c|c|c|c|c}
		\whline
		&5 	&10	&20 &30 &40 &50 &100 & 200 &300	\\
		\whline
		MAE&12.5 &12.0 &11.0 &11.2 &10.9 &11.3 &10.9 &11.4&11.2 \\
		\hline
		MSE&22.9 &20.2 &19.0 &19.8 &19.3 &20.0 &19.5 &18.5&18.7 	\\
		\whline

	\end{tabular}\label{ROIs}
\end{table*}

\begin{figure}
	\centering
	\includegraphics[width=0.48\textwidth]{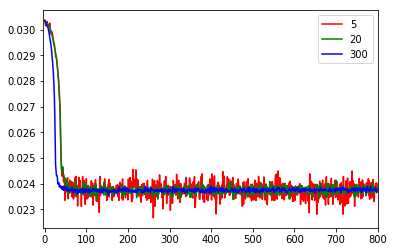}
	\caption{The curve lines of cross-entropy loss for R-HDC under the number of ROIs.}\label{Fig-loss}
\end{figure}

We note that the different numbers of ROIs influence the R-HDC's training. The more number of ROIs results in the faster convergence speed. Even so, it does not mean the model can converge to a better classification accuracy. Fig. \mbox{\ref{Fig-loss}} shows the convergence curve line of the standard 2-D cross-entropy loss for R-HDC under the number of 5, 20 and 300. In the whole experiments, the lines of 5 and 10, 20 $\sim$ 100, 200 and 300 are very close. For easier visualization, we select 5, 20 and 300 as the examples to avoid overlapping lines. From the lines, we find the convergence speed when selecting 300 ROIs is faster than others. Nevertheless, its final convergence values are almost the same as others. At the same time, we also find that more ROIs results in the more stable and smooth curve. Although R-HDC converges faster,  DME has no significant acceleration during the training process. Thus, we select the number of ROIs as $20$.

\subsection{Comparative Visualization Results with SOTA}

\label{CPCNN}

\begin{figure*}
	\centering
	\includegraphics[width=0.95\textwidth]{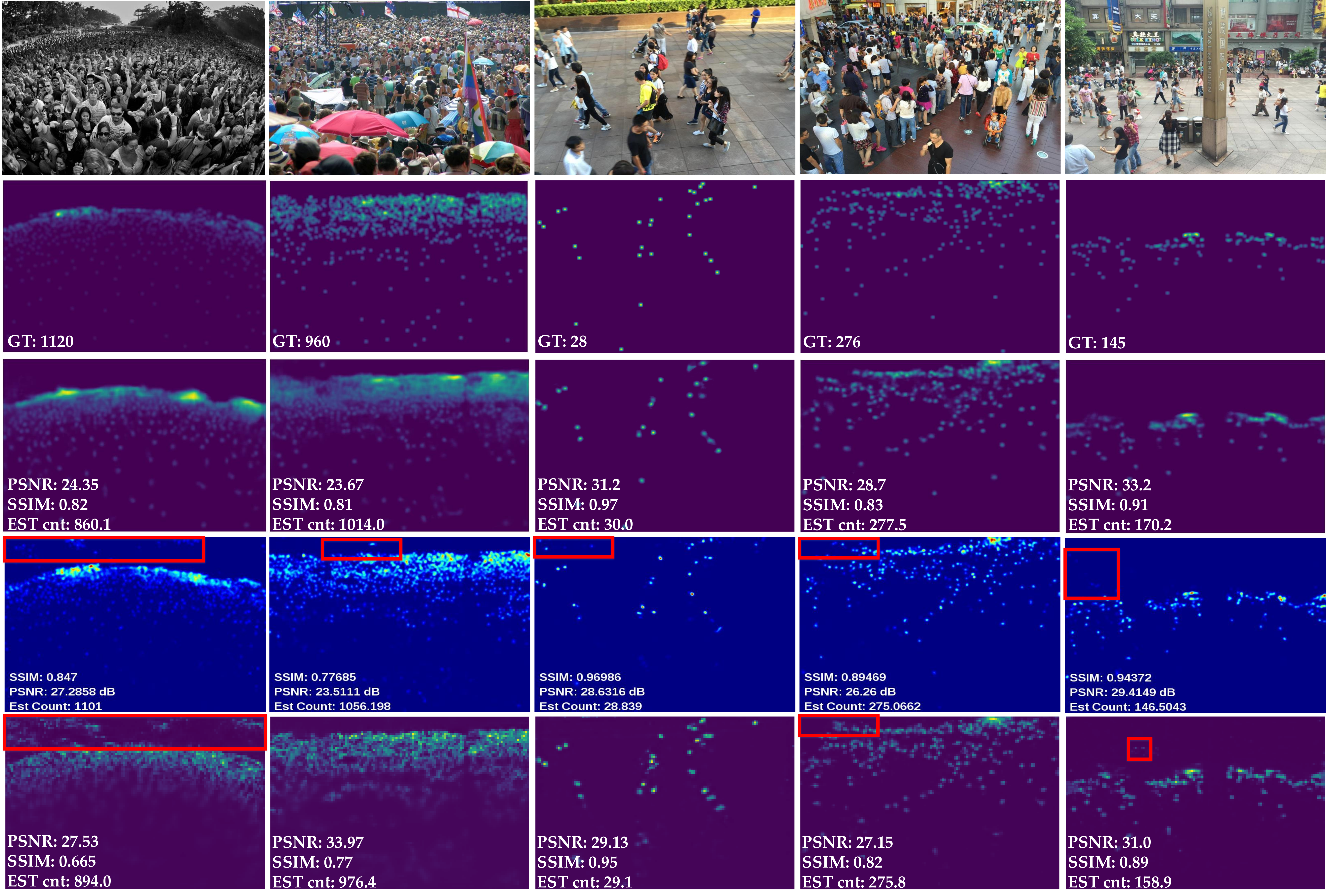}
	\caption{Comparative Visualization Results with CP-CNN and CSRNet. Row 1: Input images; Row 2: Groundtruth; Row 3: PCC Net's results; Row 4: CP-CNN's results; Row 5: CSRNet's results.}\label{Fig_viscom}
\end{figure*}

In order to show the visualization performance, especially for background regions,  we report the comparative visualization results of our proposed method,  CP-CNN \mbox{\cite{sindagi2017generating}} and CSRNet \mbox{\cite{li2018csrnet}} in Fig. \mbox{\ref{Fig_viscom}}. The first row describes the input images from Shanghai Tech A or B datasets; the second row illustrates the ground truth of density maps; and the last three rows respectively show the PCC Net's, CP-CNN's and CSRNet’s results. The red boxes in Fig. \mbox{\ref{Fig_viscom}} highlights some obvious differences between the two algorithms' results. 

From it, we find CP-CNN and CSRNet are prone to mistakenly estimating the trees or plants as the crowd. On the contrary, the proposed PCC Net performs better than CP-CNN and CSRNet in these background regions. The main reason is that PCC Net can effectively extract the discriminative features between foreground and background objects. In terms of perspective consistency in the whole image, the PCC Net can perform better than CP-CNN and CSRNet.

\subsection{The Sensibility of Segmentation Map}
In Section \mbox{\ref{Sec_FBS}}, we introduce a Fore-/Background Segmentation (FBS) into crowd counting to reduce the estimation errors. However, the original datasets do not provide the head segmentation mask. Thus, we adopt an image morphology operation to generate this mask. Here, we discuss how the setting of radius and height affect the final counting performance. Table \mbox{\ref{segmap}} lists the results under setting of radius and height on Shanghai Tech A dataset. Note that ``NoSeg'' means the model is DME+R-HDC+DULR. From the table, we find when the value is between 10  and 60 pixels, the counting results are close. When the value is less than 10 or more than 100 pixels, the performance is close to that of NoSeg. The main reason is that a too large or too small value cannot cover valid foreground and background region so that the network does not distinguish them. The learned feature maps cannot contain effective, discriminative features. When it is added to DME, there is no improvement for crowd counting.

\begin{table*}[htbp]
	
	\centering
	\caption{Performance under setting of radius and height on Shanghai Tech A dataset.}
	
	\begin{tabular}{c|c|c|c|c|c|c|c|c}
		\whline
		   &NoSeg&10  &30 &40 &\textbf{50} &60 & 100 &150	\\
		\whline
		MAE&79.3 &80.1 &74.1 &74.6 &\textbf{73.5} &73.4 &78.7 & 79.2\\
		\hline
		MSE&129.5 &128.9 &125.6 &128.3 &\textbf{124.0} &127.1 &134.9 &130.7 	\\
		\whline			
		
	\end{tabular}\label{segmap}
\end{table*}

\section{Conclusion}


In this paper, we present a multi-task Perspective Crowd Counting Network (PCC Net), which can encode hierarchical features (global, local and pixel-level features) and perspective changes for the crowd scenes. The PCC Net consists of Density Map Estimation (DME), Random High-level Density Classification (R-HDC) and Fore-/Background Segmentation (FBS). DME focuses on the learning of very local features for density map estimation. R-HDC extracts global features to predict the random image patches' coarse density labels. FBS segments the head regions and background to further remove the mistaken estimation. In addition, the DULR module is added to the DME and FBS to encode the perspective changes. Extensive experiments demonstrate that the proposed PCC Net achieves the competitive results. Especially, PCC Net significantly improves the counting performance for extremely congested crowd scenes. In the future work, we will further explore the invariant features between low and high density regions to prompt the performance for extremely congested scenes.


\bibliographystyle{IEEEtran}
\bibliography{IEEEabrv,reference}

\newpage

\begin{IEEEbiography}[{\includegraphics[width=1in,height=1.25in,clip,keepaspectratio]{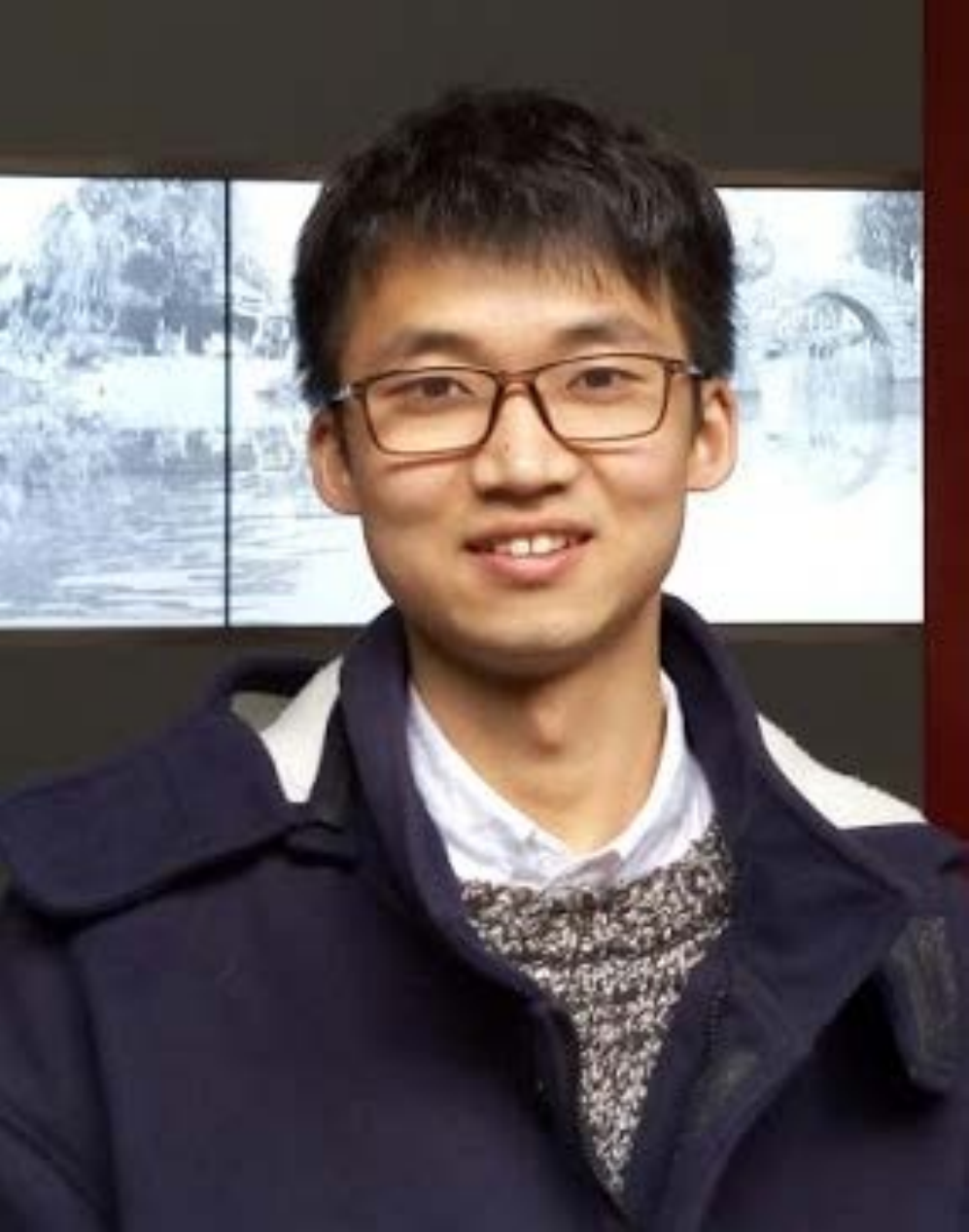}}]{Junyu Gao} received the B.E. degree in computer science and technology from the Northwestern Polytechnical University, Xi'an 710072, Shaanxi, P. R. China, in 2015. He is currently pursuing the Ph.D. degree from Center for Optical Imagery Analysis and Learning, Northwestern Polytechnical University, Xi’an, China. His research interests include computer vision and pattern recognition.
\end{IEEEbiography}

\begin{IEEEbiography}[{\includegraphics[width=1in,height=1.25in,clip,keepaspectratio]{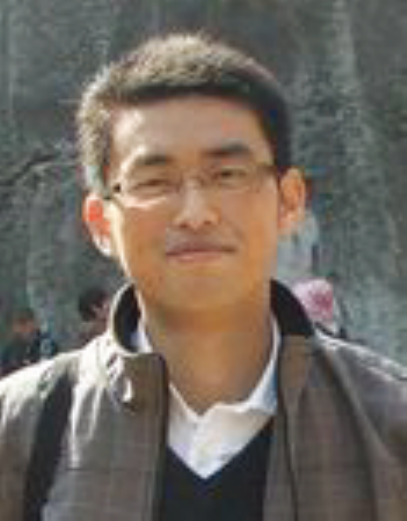}}]{Qi Wang} (M'15-SM'15) received the B.E. degree in automation and the Ph.D. degree in pattern recognition and intelligent systems from the University of Science and Technology of China, Hefei, China, in 2005  and 2010, respectively.  He is currently a Professor with the School of Computer Science and with the Center for OPTical IMagery Analysis and Learning (OPTIMAL), Northwestern Polytechnical University, Xi'an, China. His research interests include computer vision and pattern recognition.	

\end{IEEEbiography}

\begin{IEEEbiographynophoto}{Xuelong Li} (M'02-SM'07-F'12) is a full professor with the School of Computer Science and the Center for OPTical IMagery Analysis and Learning (OPTIMAL), Northwestern Polytechnical University, Xi’an 710072, Shaanxi, P. R. China.
\end{IEEEbiographynophoto}

\end{document}